# Toward Real-World IoT Security: Concept Drift-Resilient IoT Botnet Detection via Latent Space Representation Learning and Alignment

Hassan Wasswa
*School of Systems and Computing*
*University of New South Wales*
Canberra, Australia
h.wasswa@unsw.edu.au

Timothy Lynar
*School of Systems and Computing*
*University of New South Wales*
Canberra, Australia
t.lynar@unsw.edu.au

*Abstract*— Although AI-based models have achieved high accuracy in IoT threat detection, their deployment in enterprise environments is constrained by reliance on stationary datasets that fail to reflect the dynamic nature of real-world IoT NetFlow traffic, which is frequently affected by concept drift. Existing solutions typically rely on periodic classifier retraining, resulting in high computational overhead and the risk of catastrophic forgetting. To address these challenges, this paper proposes a scalable framework for adaptive IoT threat detection that eliminates the need for continuous classifier retraining. The proposed approach trains a classifier once on latent-space representations of historical traffic, while an alignment model maps incoming traffic to the learned historical latent space prior to classification, thereby preserving knowledge of previously observed attacks. To capture inter-instance relationships among attack samples, the low-dimensional latent representations are further transformed into a graph-structured format and classified using a graph neural network. Experimental evaluations on real-world heterogeneous IoT traffic datasets demonstrate that the proposed framework maintains robust detection performance under concept drift. These results highlight the framework's potential for practical deployment in dynamic and large-scale IoT environments.

## I. INTRODUCTION

The Internet of Things (IoT) has improved automation and connectivity but introduced severe security challenges, particularly from botnet attacks. Recent works have applied traditional and deep learning methods for IoT attack detection, including CNNs [1], GNNs [2], attention-based LSTMs [3], game-theoretic models [4], and LSTM-based approaches [5], demonstrating strong results on benchmark datasets like Bot-IoT [6], N-BaIoT [7], CICIoT2022 [8], and others [9–24].

However, despite their wide adoption in areas such as fault detection [25, 26], computer vision [27–29], and complex tasks including knowledge graph completion [30, 31], and aviation safety [32, 33], their application to IoT botnet detection has been hindered by the non-stationary nature of IoT attack traffic. Most existing models assume stationary traffic patterns, an assumption that is unrealistic in dynamic IoT environments. Due to evolving attacks and diverse device configurations [34], concept drift [35–39] undermines detection performance and generalization across datasets [40, 41]. Variations in attack scripts and device heterogeneity further degrade model accuracy [35].

Adaptive learning methods have been proposed [34, 40, 42], but frequent retraining causes catastrophic forgetting, model instability, and high computational cost. In addition, many of the models treat each attack instance to be independent from other instances ignoring the inter-instance relationship exhibited by IoT botnet attacks, particularly DDoS attack that originated from various zombie network nodes. To learn inter-instance relationships, graph neural network models are continuously being studied for efficient detection of IoT botnet attacks [43–45].

To address this, we propose a hybrid framework that combines Variational autoencoders, latent space representation learning and alignment [46–48], and Graph neural network (GNN) that encodes new traffic into a latent representation aligned with previously learned representations, converts it into a graph structure and uses a Graph attention (GAT) model to classify incoming traffic as either benign or malicious. In this approach, only the latent space alignment model is updated in the event of a drift, preserving classifier knowledge, reducing retraining needs, and improving detection robustness in non-stationary environments.

The rest of this paper is organised as follows. Section II presents a review of related work. This is followed by section III where a detailed description of the proposed framework is presented. Section IV provides a description of the experimental flow. Section V presents and discusses the findings of this study and finally section VI presents a conclusion of this work

## II. RELATED WORK

This section summarizes prior research in three key areas: (1) latent space alignment, (2) IoT attack detection under concept drift, and (3) GNN-based models for IoT-based attack detection.

### A. Latent Space Alignment

Study [43] proposed a 3D-VAE-based domain adaptation approach to learn a unified latent distribution for training and testing datasets, improving character recognition accuracy by 5%. However, it relied on computationally intensive 3D convolutional architectures with limited gains over 2D representations. In [44], the authors introduced a few-shot learning method to align latent spaces of non-paired image and avatar autoencoders for high-resolution avatar generation.

Work in [45] addressed the challenge of large domain gaps by training a VAE on source data and aligning target features via reconstruction with a Gaussian prior. Similarly, [46] used three VAEs to align cross-modal latent spaces (e.g., image, speech, text), overcoming limitations of shared latent representation across modalities. confirm that you have the correct template for your paper size. This template has been tailored for output on the A4 paper size. If you are using US letter-sized paper, please close this file and download the Microsoft Word, Letter file.

*B. IoT Attack Detection under Concept Drift*

In [41], the authors showed that concept drift degrades detection performance and analysed it using sliding windows and multiple datasets, but did not propose a mitigation approach. Study [34] proposed retraining on drift-causing instances while preserving prior knowledge using a Kafka-Spark-MongoDB pipeline, improving accuracy from 97.8% to 99.46%. However, frequent retraining is costly and the method was tested on limited attack variations. This work was extended in [47] for unsupervised learning but lacked concept drift validation.

The method in [35] combined PCA-based drift detection with an online DNN using Hedge weighting, evaluated on DS2OS data. While effective, evaluation on a single dataset may not reflect real-world variability. INSOMNIA [48] used semi-supervised learning with active and incremental updates to adapt to drift, reducing update latency, but faced risks of forgetting and limited generalization across datasets.

In [40], K-means-based drift detection triggered retraining in an ensemble model, which was extended in [49] with error-based and distribution-based drift detectors. Despite good results across datasets, continuous retraining remained a limitation.

*C. GNN-based models for IoT-based attack detection*

Graph-based intrusion detection has emerged as a promising direction for addressing IoT security challenges arising from heterogeneous and highly interconnected devices. In this context, [44] proposed a GAT-driven intrusion detection system that models IoT traffic as graph-structured data, demonstrating that GNNs can provide scalable and robust intrusion detection. Extending this line of work, [54] introduced a framework combining a GCN with an attention mechanism to detect IoT botnets by learning interaction patterns among devices and uncovering latent malicious behaviours.

To enhance the expressive power of graph-based models, several studies have integrated GNNs with complementary architectures. The hybrid model proposed in [55] combines a GNN with a transformer, where the GNN captures structural dependencies among nodes and edges, while the transformer models long-range contextual relationships to improve anomaly detection. Similarly, [56] introduced EGAT-LSTM, which integrates an enhanced GAT with an LSTM to jointly capture spatial and temporal characteristics of IoT traffic.

More recent work has focused on performance optimization, representation learning, and domain-specific applications of GNN-based intrusion detection. In [10], the authors investigated the impact of dimensionality reduction techniques—AE, VAE, and PCA—on GAT-based IoT botnet detection, showing that VAE-based compression of the CICIoT2022 feature space yielded superior results when combined with kNN graph construction. A broader comparison conducted in [13] revealed that while VAE-GAT outperformed VAE-GCN on the N-BaIoT dataset, both models were surpassed by non-GNN architectures, indicating the need for further refinement. Beyond traditional IoT networks, [57] proposed AJSAGE, an attention-augmented GraphSAGE model that improves the detection of sophisticated anomalies in graph-based network traffic.

This work differs by introducing a latent space alignment strategy that avoids frequent classifier retraining. Incoming traffic is encoded and aligned with previously learned representations, updating only the alignment model. This preserves classifier knowledge, prevents catastrophic forgetting, and reduces retraining costs in non-stationary IoT environments.

## III. PROPOSED APPROACH

The proposed approach consists of four main steps as illustrated in Fig. 1 (coded in respective colours). In the first step, a Variational Autoencoder (VAE) is trained using historically observed IoT botnet traffic. Its encoder, $E_H$, is used to project the high-dimensional historical data into a low-dimensional latent space.

In the second step, the low dimensional dataset from historical data is converted into a graph structured dataset using the k-Nearest Neighbour algorithm with n_neighbors set to 3 and using Euclidean distance as the metric. The resulting graph dataset is then used to train a GAT model for node (traffic) classification.

In the third step, a VAE is trained on new attack traffic collected from a heterogeneous network and exhibiting concept drifts. Using the encoder component, $E_S$, the high dimensional traffic is transform into a low dimensional dataset.

Step 4 involves training the alignment model. The model is trained to align the latent vector from new attack traffic with the latent vectors of historical dataset. This allows the model trained on historical attack traffic to efficient detect traffic from heterogeneous attack sources, characterised with concept drift, with retraining.

During deployment, incoming traffic is passed through the encoder $E_S$, aligned with the historical latent space, and then sent to the detection model $C_H$ for classification. Notably, the alignment model can be updated independently, without retraining $C_H$, thereby preserving historically learned patterns and reducing the cost associated with frequent model retraining.

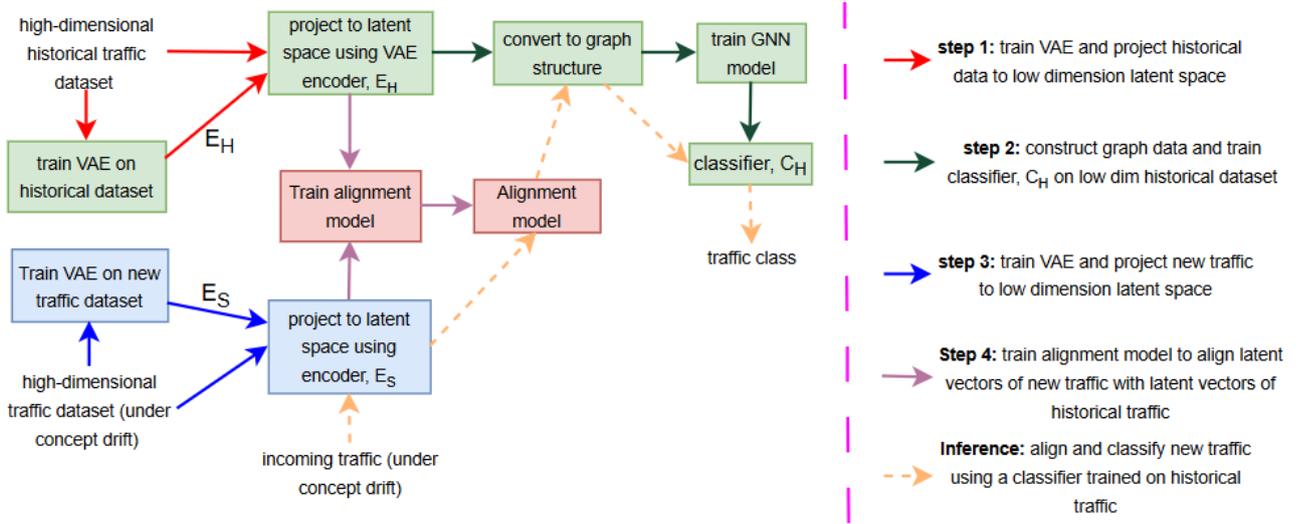

Figure 1: Proposed framework

## A. Alignment model

In this study, the alignment model is implemented as a Multi-Layer Perceptron (MLP), which learns a transformation that aligns the latent space derived from current IoT traffic with the latent space obtained from historical data. This alignment is achieved by minimizing the Wasserstein Distance, ($W_d$)—defined in Eq. 1— between the two latent space distributions.

$$W_d = (\|\mu_H - \mu_S\|^2 + \|\sigma_H - \sigma_S\|^2)^{1/2} \qquad (1)$$

where:

$\mu_H$ and $\sigma_H$ denote the mean and standard deviation, respectively, of the latent space distribution learned from historical data.

$\mu_S$ and $\sigma_S$ denote the mean and standard deviation, respectively, of the latent space distribution learned from new data.

## B. Datasets

This study addressed a binary classification task to distinguish between "Normal" and "Attack" traffic using two publicly available datasets: ACI-IoT-2023 [50], and the IoT Network Intrusion Dataset (IoT-NID) [51].

### 1) ACI-IoT-2023 dataset

The ACI-IoT-2023 [50] dataset was produced in a lab mimicking a real home IoT setup at the Army Cyber Institute. Devices were managed via the "Home Assistant" platform, with both wired and wireless network segments monitored. Over five days, benign and malicious traffic (e.g., reconnaissance, DDoS, spoofing, and brute force attacks) was recorded, offering a realistic testbed for IoT security research.

### 2) IoT Network Intrusion Dataset (IoT-NID)

The IoT-NID [51] dataset simulates various attacks in a wireless IoT environment comprising smart devices (e.g., speaker, camera) and mobile endpoints. Traffic was captured using a monitoring-mode wireless adapter, resulting in 42 packet capture files. It includes benign activity and attacks like ARP spoofing, SYN flooding, reconnaissance scans, and Mirai-based assaults such as UDP flooding and HTTP brute force.

## IV. EXPERIMENTS

### A. Data preprocessing

During data cleaning, the datasets were refined by discarding anomalies, repeated entries, incomplete data, and incorrect values. Columns like "Flow ID", "Src IP", "Dst IP", and "Timestamp" were excluded, as they did not contribute meaningful patterns for the learning process and could introduce bias.

### B. Experiments for Concept drift illustration

To assess the impact of concept drift on IoT detection performance, a classifier $C_H$ was trained and evaluated on the ACI-IoT-2023 dataset, following the procedure illustrated in steps 1 and 2 of Fig. 1. Using the resulting encoder model $E_H$, test samples from the IoT-NID2024 dataset were projected into a low-dimensional representation and then passed to $C_H$ for classification without alignment. The experiment was then repeated with the roles of the datasets reversed: the IoT-NID2024 dataset was treated as the historical dataset, and the ACI-IoT-2023 dataset was treated as the current dataset. For all experiments, the latent space dimension was set to 8.

### C. Experiments for latent space alignment

The ACI-IoT-2023 dataset was treated as historically observed IoT attack traffic, while the IoTNID2024 dataset represented current IoT attack traffic. A Variational

Autoencoder (VAE) was trained on the ACI-IoT-2023 dataset, and its encoder was used to project the high-dimensional data into a low-dimensional latent space (Figure 1: step 1). The dataset was then converted into a graph dataset and classifier trained and evaluated using low-dimensional representation test samples of the ACI-IoT-2023 dataset (Figure 1: step 2).

To enable alignment, a second VAE was trained on the IoT-NID2024 dataset, and train samples from the IoT-NID2024 dataset were projected into the low dimension space using $E_S$ model (Figure 1: step 3). The latent space alignment model was then trained and used to map the latent space learned from the IoT-NID2024 dataset to that of the ACI-IoT-2023 dataset. Finally, high-dimensional test samples from the IoTNID2024 dataset were projected into a low-dimensional latent space using the encoder $E_S$, aligned with the historical latent space representations, converted into a graph data structure and subsequently fed into the classifier $C_H$ for classification.

### D. Evaluation metrics

The study employed metrics, including Accuracy, Precision, and Recall to evaluate the effectiveness of the proposed method in identifying different attack categories within the experimental datasets. These metrics report the weighted average performance of the detection model. In addition, the confusion matrix was used to show distribution of attack traffic across the benign and attack classes.

Table 1: Summary of the evaluation metrics used

| Metric used | Formula |
| --- | --- |
| Accuracy | (TP + TN) / (TP + FP + TN + FN) |
| Precision | TP / (TP + FP) |
| Recall | TP / (TP + TN) |
| F1-score | (2 ∗ precision ∗ recall) / (precision+ recall) |

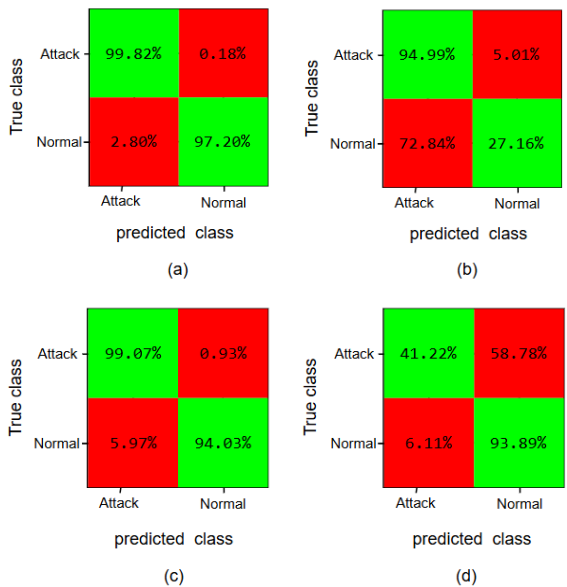

Figure 2: Impact of concept drift, model classifier: (a) trained and tested on ACI-IoT-2023; (b) trained on ACI-IoT-2023 and tested on IoT-NID2024 test samples; (c) trained and tested on IoT-NID2024 (d) trained on IoT-NID2024 and tested on ACI-IoT-2023 test samples.

## V. RESULTS AND DISCUSSION

The results from the experiments conducted in Subsection 4.2 are presented in Fig. 3. It is evident that a model trained and tested on traffic instances from a homogeneous network configuration achieves impressive results—highlighted in green when both training and testing instances are drawn from the ACI-IoT-2023 dataset, and in pink when both are from the IoTNID2024 dataset. Additionally, an accuracy of 98.46% was recorded when the train and test instances came from the ACI-IoT network while 97.93% accuracy was recorded when the train and test instances came from the IoT-NID network.

Conversely, the model's performance degrades significantly when the training and testing traffic originate from heterogeneous IoT network configurations. A model trained on traffic instances from the ACI-IoT network recorded an accuracy of 59.82% when evaluated on test instances from the IoT-NID network. Similarly, a model trained on traffic instances from the IoT-NID network recorded an accuracy of 53.11% when evaluated on test instances from the ACI-IoT network. This indicates that traffic from different IoT networks introduces concept drift, which negatively impacts the classifier's performance. Such drift is a common challenge in today's highly dynamic IoT attack landscape. The findings are further visualized using confusion matrix plots in Fig. 2

| | | Test traffic source | | | | | |
| --- | --- | --- | --- | --- | --- | --- | --- |
| | | Precision (%) | | Recall (%) | | F1-Score (%) | |
| | | ACI-IoT | IoT-NID | ACI-IoT | IoT-NID | ACI-IoT | IoT-NID |
| Train traffic | ACI-IoT | 98.51 | 70.65 | 98.46 | 59.82 | 98.46 | 54.83 |
| | IoT-NID | 81.41 | 97.92 | 53.11 | 97.93 | 55.35 | 97.92 |

Figure 3: Impact of Concept drift using IoT botnet benchmark datasets

### A. Results from latent space alignment

In the experiments presented in Subsection 4.3, the low-dimensional representations of the IoT-NID2024 dataset are aligned with the latent space learned from the ACI-IoT-2023 dataset. As a result, the model $C_H$, which was trained on the ACI-IoT-2023 dataset, demonstrates significantly improved detection performance across all three evaluation metrics: accuracy increases from 59.82% to 96.56%, precision from 70.65% to 96.57%, recall from 59.82% to 96.56%, while the F1-score increased from 54.83% to 96.56%. The distribution of the IoTNID2024 test instances is further visualized by the confusion matrix in Fig. 4.

Figure 4: After latent space alignment: classifier trained on ACI-IoT-2023 and tested on IoT-NID2024

## VI. Conclusion

This research presents a framework for detecting IoT botnet attacks in environments where network conditions and attack patterns frequently evolve. The approach leverages a Variational Autoencoder to extract meaningful latent features from past data, builds a graph structured embeddings and trains a GAT classifier on these embeddings. A key innovation is the alignment mechanism that adjusts incoming data affected by concept drift to fit within the previously learned latent space, allowing accurate classification without retraining the main detection model. The method was evaluated using two real-world datasets and showed consistent improvements in handling nonstationary data through latent space alignment. However, the current framework lacks integrated concept drift detection tools. As future work, the researchers aim to identify and incorporate the best performing drift detection techniques and to develop adaptive strategies for updating the alignment model, improving the system's resilience over time.

## References

[1] B. Ahmad, Z. Wu, Y. Huang, S. U. Rehman, Enhancing the security in iot and iiot networks: An intrusion detection scheme leveraging deep transfer learning, Knowledge-Based Systems 305 (2024) 112614.

[2] H. Nguyen, R. Kashef, Ts-ids: Traffic-aware self-supervised learning for iot network intrusion detection, Knowledge-Based Systems 279 (2023) 110966.

[3] K. Swathi, G. H. Bindu, An automated intrusion detection system in iot system using attention based deep bidirectional sparse auto encoder model, Knowledge-Based Systems 305 (2024) 112633.

[4] S. Kumar, A. kumar Keshri, An effective ddos attack mitigation strategy for iot using an optimization-based adaptive security model, Knowledge-Based Systems 299 (2024) 112052.

[5] J. Zhao, M. Shao, H. Wang, X. Yu, B. Li, X. Liu, Cyber threat prediction using dynamic heterogeneous graph learning, Knowledge-Based Systems 240 (2022) 108086.

[6] N. Koroniotis, N. Moustafa, E. Sitnikova, B. Turnbull, Towards the development of realistic botnet dataset in the internet of things for network forensic analytics: Bot-iot dataset, Future Generation Computer Systems 100 (2019) 779–796.

[7] Y. Meidan, M. Bohadana, Y. Mathov, Y. Mirsky, A. Shabtai, D. Breitenbacher, Y. Elovici, N-baiot—network-based detection of iot botnet attacks using deep autoencoders, IEEE Pervasive Computing 17 (3) (2018) 12–22.

[8] S. Dadkhah, H. Mahdikhani, P. K. Danso, A. Zohourian, K. A. Truong, A. A. Ghorbani, Towards the development of a realistic multidimensional iot profiling dataset, in: 2022 19th Annual International Conference on Privacy, Security & Trust (PST), IEEE, 2022, pp. 1–11.

[9] M. Al-Fawa'Reh, J. Abu-Khalaf, P. Szewczyk, J. J. Kang, Malbot-drl: Malware botnet detection using deep reinforcement learning in iot networks, IEEE Internet of Things Journal 11 (6) (2024) 9610 – 9629. doi:10.1109/JIOT.2023.3324053.

[10] H. Wasswa, H. Abbass, T. Lynar, Graph attention neural network for botnet detection: Evaluating autoencoder, vae and pca-based dimension reduction, arXiv preprint arXiv:2505.17357 (2025).

[11] H. Nandanwar, R. Katarya, Tl-bilstm iot: transfer learning model for prediction of intrusion detection system in iot environment, International Journal of Information Security 23 (2) (2024) 1251 – 1277. doi:10.1007/s10207-023-00787-8.

[12] A. Nazir, J. He, N. Zhu, S. S. Qureshi, S. U. Qureshi, F. Ullah, A. Wajahat, M. S. Pathan, A deep learning-based novel hybrid cnn-lstm architecture for efficient detection of threats in the iot ecosystem, Ain Shams Engineering Journal 15 (7) (2024). doi:10.1016/j.asej.2024.102777.

[13] H. Wasswa, H. Abbass, T. Lynar, Are gnns worth the effort for iot botnet detection? a comparative study of vae-gnn vs. vit-mlp and vae-mlp approaches, arXiv preprint arXiv:2505.17363 (2025).

[14] F. Wahab, A. Shah, I. Khan, B. Ali, M. Adnan, An sdn-based hybrid-dl-driven cognitive intrusion detection system for iot ecosystem, Computers and Electrical Engineering 119 (2024). doi:10.1016/j.compeleceng.2024.109545.

[15] H. Wasswa, T. Lynar, H. Abbass, Enhancing iot-botnet detection using variational auto-encoder and cost-sensitive learning: A deep learning approach for imbalanced datasets, in: 2023 IEEE Region 10 Symposium, TENSYMP 2023, IEEE, 2023, pp. 1–6. doi:10.1109/TENSYMP55890.2023.10223613.

[16] N. Abdalgawad, A. Sajun, Y. Kaddoura, I. A. Zualkernan, F. Aloul, Generative deep learning to detect cyberattacks for the iot-23 dataset, IEEE Access 10 (2021) 6430–6441.

[17] H. Wasswa, A. Nanyonga, T. Lynar, Impact of latent space dimension on iot botnet detection performance: Vae-encoder versus vit-encoder, in: 2024 3rd International Conference for Innovation in Technology (INOCON), 2024, pp. 1–6. doi:10.1109/INOCON60754.2024.10511431.

[18] H. Wasswa, T. Lynar, A. Nanyonga, H. Abbass, Iot botnet detection: Application of vision transformer to classification of network flow traffic, in: 2023 Global Conference on Information Technologies and Communications (GCITC), 2023, pp. 1–6. doi:10.1109/GCITC60406.2023.10426522.

[19] N. Abdalgawad, A. Sajun, Y. Kaddoura, I. A. Zualkernan, F. Aloul, Generative deep learning to detect cyberattacks for the iot-23 dataset, IEEE Access 10 (2022) 6430–6441. doi:10.1109/ACCESS.2021.3140015.

[20] A. R. Gad, M. Haggag, A. A. Nashat, T. M. Barakat, A distributed intrusion detection system using machine learning for iot based on ton-iot dataset, International Journal of Advanced Computer Science and Applications 13 (6) (2022).

[21] S. Nomm, Using medbiot dataset to build effective machine learning-based iot botnet detection systems, in: Information Systems Security and Privacy: 6th International Conference, ICISSP 2020, Valletta, Malta, February 25–27, 2020, Revised Selected Papers, Springer, 2022, p. 222.

[22] S. Saraniya, M. Sowmiya, B. Kalpana, M. Karthi, Securing networks: Unleashing the power of the ft-transformer for intrusion detection, in: 2024 International Conference on Computer, Electrical & Communication Engineering (ICCECE), IEEE, 2024, pp. 1–7.

[23] H. Wasswa, H. A. Abbass, T. Lynar, Resdnvit: A hybrid architecture for netflow-based attack detection using a residual dense network and vision transformer, Expert Systems with Applications (2025) 127504.

[24] I. Zakariyya, H. Kalutarage, M. O. Al-Kadri, Robust, effective and resource efficient deep neural network for intrusion detection in iot networks, in: Proceedings of the 8th ACM on Cyber-Physical System Security Workshop, 2022, pp. 41–51.

[25] W. Sun, X. Shi, S. He, W. Xiong, H. Chen, B. Huang, A conditional invertible neural network-based fault detection, IEEE Transactions on Control Systems Technology (2025).

[26] P. Wu, E. Tian, H. Tao, Y. Chen, Data-driven spiking neural networks for intelligent fault detection in vehicle lithium-ion battery systems, Engineering Applications of Artificial Intelligence 141 (2025) 109756.

[27] G. Casqueiro, S. E. Arefin, T. A. Heya, A. Serwadda, H. Wasswa, Weaponizing iot sensors: When table choice poses a security vulnerability, in: 2022 IEEE 4th International Conference on Trust,

[28] H. Wasswa, A. Serwadda, The proof is in the glare: On the privacy risk posed by eyeglasses in video calls, in: Proceedings of the 2022 ACM on International Workshop on Security and Privacy Analytics, 2022, pp. 46–54.

[29] M. Yuzhan, H. Abdullah Jalab, W. Hassan, D. Fan, M. Minjin, Recaptured image forensics based on image illumination and texture features, in: Proceedings of the 2020 4th International Conference on Video and Image Processing, 2020, pp. 93–97.

[30] D. Yuan, S. Zhou, X. Chen, D. Wang, K. Liang, X. Liu, J. Huang, Knowledge graph completion with relation-aware anchor enhancement, in: Proceedings of the AAAI Conference on Artificial Intelligence, Vol. 39, 2025, pp. 15239–15247.

[31] R. Yang, J. Zhu, J. Man, H. Liu, L. Fang, Y. Zhou, Gs-kgc: A generative subgraph-based framework for knowledge graph completion with large language models, Information Fusion 117 (2025) 102868.

[32] A. Nanyonga, H. Wasswa, K. Joiner, U. Turhan, G. Wild, Explainable supervised learning models for aviation predictions in australia, Aerospace 12 (3) (2025) 223.

[33] A. Nanyonga, H. Wasswa, K. Joiner, U. Turhan, G. Wild, A multi-head attention-based transformer model for predicting causes in aviation incidents, Modelling 6 (2) (2025) 27.

[34] S. Saravanan, U. M. Balasubramanian, An adaptive scalable data pipeline for multiclass attack classification in large-scale iot networks, Big Data Mining and Analytics 7 (2) (2024) 500–511.

[35] O. A. Wahab, Intrusion detection in the iot under data and concept drifts: Online deep learning approach, IEEE Internet of Things Journal 9 (20) (2022) 19706–19716.

[36] Y. Wen, X. Liu, H. Yu, Adaptive tree-like neural network: Overcoming catastrophic forgetting to classify streaming data with concept drifts, Knowledge-Based Systems 293 (2024) 111636.

[37] S. Cai, Y. Zhao, Y. Hu, J. Wu, J. Wu, G. Zhang, C. Zhao, R. N. A. Sosu, Cd-btmse: A concept drift detection model based on bidirectional temporal convolutional network and multi-stacking ensemble learning, Knowledge-Based Systems 294 (2024) 111681.

[38] Y. Sun, J. Mi, C. Jin, Entropy-based concept drift detection in information systems, Knowledge-Based Systems 290 (2024) 111596.

[39] X. Zheng, P. Li, X. Hu, K. Yu, Semi-supervised classification on data streams with recurring concept drift and concept evolution, Knowledge-Based Systems 215 (2021) 106749.

[40] M. Jain, G. Kaur, Distributed anomaly detection using concept drift detection based hybrid ensemble techniques in streamed network data, Cluster Computing 24 (3) (2021) 2099–2114.

[41] H. Qiao, B. Novikov, J. O. Blech, Concept drift analysis by dynamic residual projection for effectively detecting botnet cyber-attacks in iot scenarios, IEEE Transactions on Industrial Informatics 18 (6) (2021) 3692–3701.

[42] Y. K. Beshah, S. L. Abebe, H. M. Melaku, Drift adaptive online ddos attack detection framework for iot system, Electronics 13 (6) (2024) 1004.

[43] L. Lin, Q. Zhong, J. Qiu, Z. Liang, E-gracl: an iot intrusion detection system based on graph neural networks, The Journal of Supercomputing 81 (1) (2025) 42.

[44] A. S. Ahanger, S. M. Khan, F. Masoodi, A. O. Salau, Advanced intrusion detection in internet of things using graph attention networks, Scientific Reports 15 (1) (2025) 9831.

[45] P. H. Do, T. L. Tran, A. A. Ateya, T. D. Le, et al., Investigating the effectiveness of different gnn models for iot-healthcare systems botnet traffic classification, in: Secure Health, CRC Press, 2025, pp. 104–123.

[46] M. L. De Guevara, J. Echevarria, Y. Li, Y. Hold-Geoffroy, C. Smith, D. Ito, Cross-modal latent space alignment for image to avatar translation, in: Proceedings of the IEEE/CVF International Conference on Computer Vision, 2023, pp. 520–529.

[47] H. Wasswa, H. A. Abbass, T. Lynar, Latent space alignment for robust detection of iot botnet attacks in non-stationary environments, Knowledge-Based Systems (2025) 114749.

[48] C. Wu, Z. Du, J. Gan, G. Zhang, Y. Wang, X. Wang, P. Wu, F. Zhou, Cross-domain character recognition through latent space alignment, in: Proceedings of the 5th International Conference on Big Data Technologies, 2022, pp. 266–272.

[49] J. Wang, J. Chen, J. Lin, L. Sigal, C. W. de Silva, Discriminative feature alignment: Improving transferability of unsupervised domain adaptation by gaussian-guided latent alignment, Pattern Recognition 116 (2021) 107943.

[50] T. Theodoridis, T. Chatzis, V. Solachidis, K. Dimitropoulos, P. Daras, Cross-modal variational alignment of latent spaces, in: Proceedings of the IEEE/CVF conference on computer vision and pattern recognition workshops, 2020, pp. 960–961.

[51] S. Saravanan, U. M. Balasubramanian, Uasdac: An unsupervised adaptive scalable ddos attack classification in large-scale iot network under concept drift, IEEE Access (2024).

[52] G. Andresini, F. Pendlebury, F. Pierazzi, C. Loglisci, A. Appice, L. Cavallaro, Insomnia: Towards concept-drift robustness in network intrusion detection, in: Proceedings of the 14th ACM workshop on artificial intelligence and security, 2021, pp. 111–122.

[53] M. Jain, G. Kaur, V. Saxena, A k-means clustering and svm based hybrid concept drift detection technique for network anomaly detection, Expert Systems with Applications 193 (2022) 116510.

[54] J. Kumar, et al., Grma-cnn: Integrating spatial-spectral layers with modified attention for botnet detection using graph convolution for securing networks., International Journal of Intelligent Engineering & Systems 18 (1) (2025).

[55] H. Zhang, T. Cao, A hybrid approach to network intrusion detection based on graph neural networks and transformer architectures, in: 2024 14th International Conference on Information Science and Technology (ICIST), IEEE, 2024, pp. 574–582.

[56] L. Zhang, L. Tan, H. Shi, H. Sun, W. Zhang, Malicious traffic classification for iot based on graph attention network and long short-term memory network, in: 2023 24st Asia-Pacific Network Operations and Management Symposium (APNOMS), IEEE, 2023, pp. 54–59.

[57] L. Xu, Z. Zhao, D. Zhao, X. Li, X. Lu, D. Yan, Ajsage: A intrusion detection scheme based on jump-knowledge connection to graphsage, Computers & Security 150 (2025) 104263.

[58] N. Bastian, D. Bierbrauer, M. McKenzie, E. Nack, Aci iot network traffic dataset 2023 (2023). doi:10.21227/qacj-3x32. URL https://dx.doi.org/10.21227/qacj-3x32

[59] H. Kang, D. H. Ahn, G. M. Lee, J. D. Yoo, K. H. Park, H. K. Kim, Iot network intrusion dataset (2019). doi:10.21227/q70p-q449. URL https://dx.doi.org/10.21227/q70p-q449